\documentclass[10pt,twocolumn,letterpaper]{article}

\usepackage{cvpr}
\usepackage{times}
\usepackage{epsfig}
\usepackage{graphicx}
\usepackage{amsmath}
\usepackage{amssymb}

\usepackage{color}
\usepackage{amsmath}
\usepackage{amsfonts}
\usepackage{textcomp}
\definecolor{pink}{rgb}{0.858, 0.188, 0.478}

\cvprfinalcopy 

\def\cvprPaperID{1124} 

\ifcvprfinal\fi
\begin{document}

\title{Feature-metric Registration: A Fast Semi-supervised Approach for Robust Point Cloud Registration without Correspondences}

\author{Xiaoshui Huang\\
The University of Sydney\\
Sydney, Australia\\
{\tt\small Xiaoshui.Huang@sydney.edu.au}
\and
Guofeng Mei, Jian Zhang\\
University of Technology Sydney\\
Sydney, Australia\\
{\tt\small {Guofeng.Mei@student.uts.edu.au, Jian.Zhang@uts.edu.au}}
}

\maketitle

\begin{abstract}

We present a fast feature-metric point cloud registration framework, which enforces the optimisation of registration by minimising a feature-metric projection error without correspondences. The advantage of the feature-metric projection error is robust to noise, outliers and density difference in contrast to the geometric projection error. Besides, minimising the feature-metric projection error does not need to search the correspondences so that the optimisation speed is fast. The principle behind the proposed method is that the feature difference is smallest if point clouds are aligned very well. We train the proposed method in a semi-supervised or unsupervised approach, which requires limited or no registration label data. Experiments demonstrate our method obtains higher accuracy and robustness than the state-of-the-art methods. Besides, experimental results show that the proposed method can handle significant noise and density difference, and solve both same-source and cross-source point cloud registration.  
\end{abstract}

\section{Introduction}
Point cloud registration is a process of transforming different scans of the same 3D scene or object into one coordinate system. Most of the state-of-the-art registration methods \cite{huang2017systematic,gao2019filterreg,gojcic2019perfect} minimize a geometric {(point coordinate-based)} projection error through two processes: correspondences searching and transformation estimation. These two processes alternatively conduct until the geometric projection error is minimum. We strive to solve the point cloud registration because the registration is critical for many tasks such as robotics vision and augment reality. 

\begin{figure}[t]
	\includegraphics[width=\linewidth]{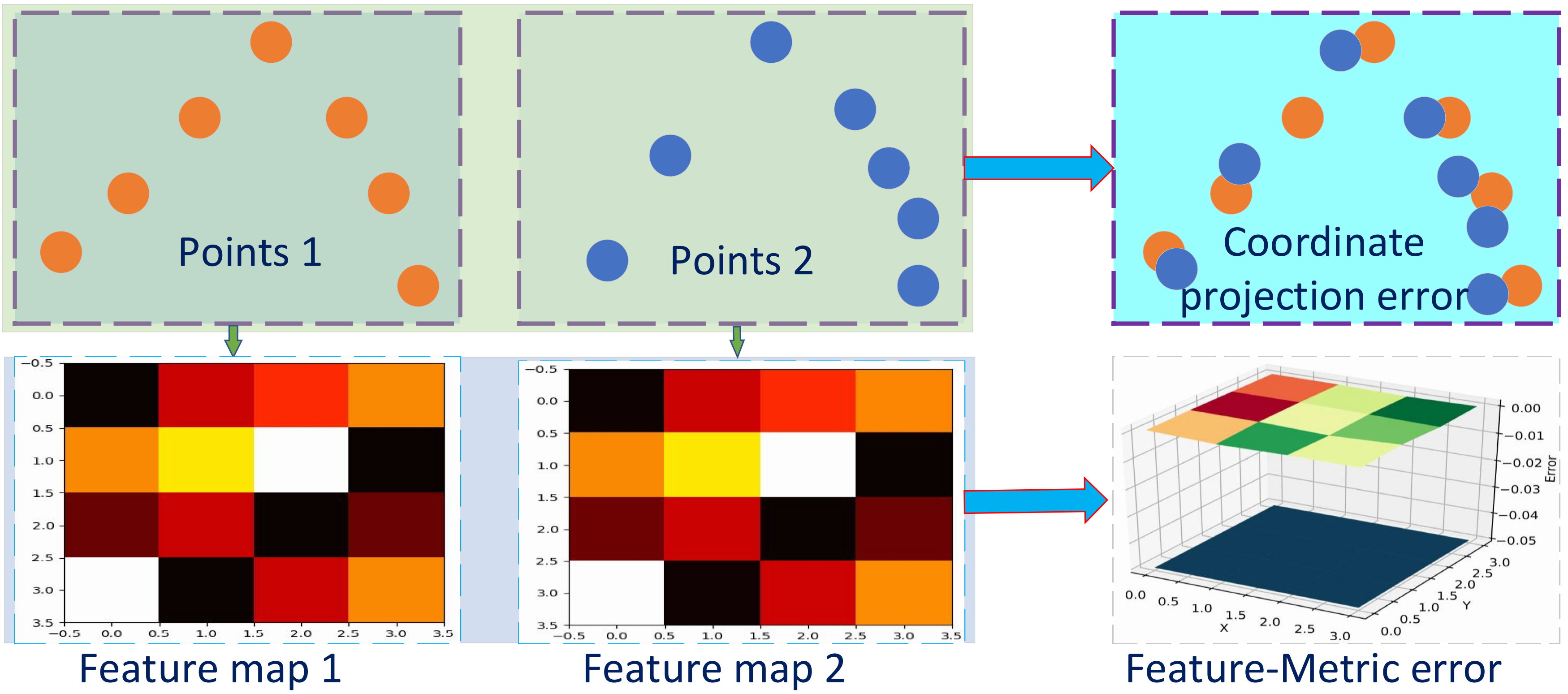}
	\caption{The difference between the coordinate-based registration (top) and feature-metric registration(bottom). The top shows the best alignment can never be achieved since this alignment has no point-point correspondences. The bottom feature-metric registration can achieve the best alignment since it optimizes the feature difference without correspondence and the feature difference is minimum in the best alignment. }
	\label{f1}
\end{figure}

In the point cloud registration task, looking at Figure \ref{f1}, the best alignment of point 1 and point 2 is the third column (manually aligned) since the geometric projection error is minimum. However, such best alignment can hardly achieve by using the existing two-process methods since there are no point-point correspondences. Even though the soft correspondence \cite{gold1998new} provides an alternative way by using weighted point-point correspondences, the behind theory still relies on the point-point correspondences, and it is tricky to decide the weights.  Besides, the noise, outliers and density differences existed in the point clouds will make the registration problem more challenging since the point-point correspondences are impaired during the acquisition process \cite{huang2017systematic}.

There are plenty of literature focus on solving point cloud registration problem \cite{yang2015go,le2019sdrsac,3dmatch,huang2017systematic,Yang_2019_CVPR,gojcic2019perfect}. Roughly, they could generally be divided into two categories: global and local methods. Global methods \cite{yang2015go,le2019sdrsac} usually estimate the up-bound and low-bound of the objective function. In contrast,  the local methods \cite{3dmatch,huang2017systematic,Yang_2019_CVPR,gojcic2019perfect} always care about local alignment. A fundamental block of the existing methods is the iterative computation of correspondences and transformation. Present implementations of this fundamental block have some inherent limitations. (1) They solve the registration tasks by minimizing a geometric projection error. This error metric is sensitive to noise and outliers since the registration relies on point-point correspondences. (2) The pre-trained descriptor training is an independent process to the registration framework. This separated training process can only learn two points which are matched or not matched based on their local regions. This match/non-match descriptor may not be helpful or work for the registration task, especially when the data has noise and outliers. (3) The existing learning-based registration methods \cite{aoki2019pointnetlk,gojcic2019perfect} rely on huge registration label data which makes the algorithms not practical since the 3D registration label is labour-consuming. These observations indicate that most of the current research work is focusing on optimization strategies or pre-trained match descriptors or supervised learning. Few pieces of research focus on the semi-supervised or unsupervised techniques to smartly transferring the registration ability acquired from aligning one pair to another without searching correspondences. 

In this paper, a feature-metric point cloud registration framework is proposed to solve the registration problem by minimizing the projection error on a feature space without the need of searching correspondence. The principle of the proposed method is that the feature difference of two-point clouds should be smallest if they align well. In particular, the proposed method contains two modules: encoder and multi-task branches. The encoder module learns to extract feature for input point clouds. For the multi-task branches module, the first branch is an encoder-decoder neural network which is designed to train the encoder module in an unsupervised manner. The goal of this branch is that the features of two copies of the same point cloud should be different if they have a spatial transformation change and the features must be the same if the transformation change is removed. The second branch is a feature-metric registration which estimates a transformation matrix between two input point clouds by minimizing the feature difference.  

The contributions of this work are mainly two aspects:   
\begin{itemize}
	\item (1) A feature-metric point cloud registration framework is proposed to solve the registration without searching correspondences. 
	\item (2) A semi-supervised approach is proposed to train the registration framework. This solution enables the feature network to learn to extract a distinctive feature for registration with limited or without registration label data.
\end{itemize}

\section{Related works}

\subsection{Optimisation-based registration methods}
Before deep learning is prevalent, most of point cloud registration methods \cite{yang2019polynomial,le2019sdrsac,pomerleau2015review,cheng2018registration} use two-stage optimisation-based algorithms: correspondence searching and transformation estimation. Most of these methods focus their research on estimating the transformation under different variations, such as noise and outliers. \cite{yang2019polynomial} uses a truncated least-squares cost to solve the extreme outliers problem in polynomial time. \cite{le2019sdrsac} claims that descriptor-based correspondence estimation becomes unreliable in noisy and contaminated setting, and proposes a randomized algorithm for robust point cloud registration without correspondences estimation.  \cite{gao2019filterreg} proposes a probabilistic registration method to formulate the E step as a filtering problem and solve it using advances in efficient Gaussian filters. \cite{pomerleau2015review,cheng2018registration} provide very good review about point cloud registration algorithms. A discussion of full literature of these methods is beyond the scope of this paper.  Although there have endured a great development on optimisation-based registration, most of the existing methods use coordinate-based Euclidean distance metric to evaluate the projection error. The limitation of the coordinate-based Euclidean distance metric is sensitive to noise and outliers. 

\subsection{Feature learning methods for registration}
Different to the classical optimization-based registration methods, numerous recent registration methods \cite{3dmatch,deng2018ppfnet,gojcic2019perfect} are tried to use the deep neural network  to learn a robust descriptor. The feature is then used to select $K$ pairs of point correspondences. Finally, the RANSAC is applied to reject the correspondence outliers and solve the point cloud registration problem. \cite{3dmatch} uses AlexNet to learn a 3D feature from an RGB-D dataset.  \cite{deng2018ppfnet} proposes a local PPF feature by using the distribution of neighbour points and then input into the network for deep feature learning. \cite{gojcic2019perfect} proposes a rotation-invariant hand-craft feature and input into a deep neural network for feature learning. All these methods are using deep learning as a feature extraction tool. Their goals are developing sophisticated strategies to learn a robust feature so that the robust correspondences can be estimated. The major limitation of this kind of methods is that they use a separated training process to learn a stand-alone feature extraction network. The learned feature network is to determine point-point matching. Different to this kind of methods, we use the neural network to learn the feature for the final registration directly and use the feature-metric projection error to estimate the transformation relationship.

\begin{figure*}[ht]
	\includegraphics[width=\linewidth]{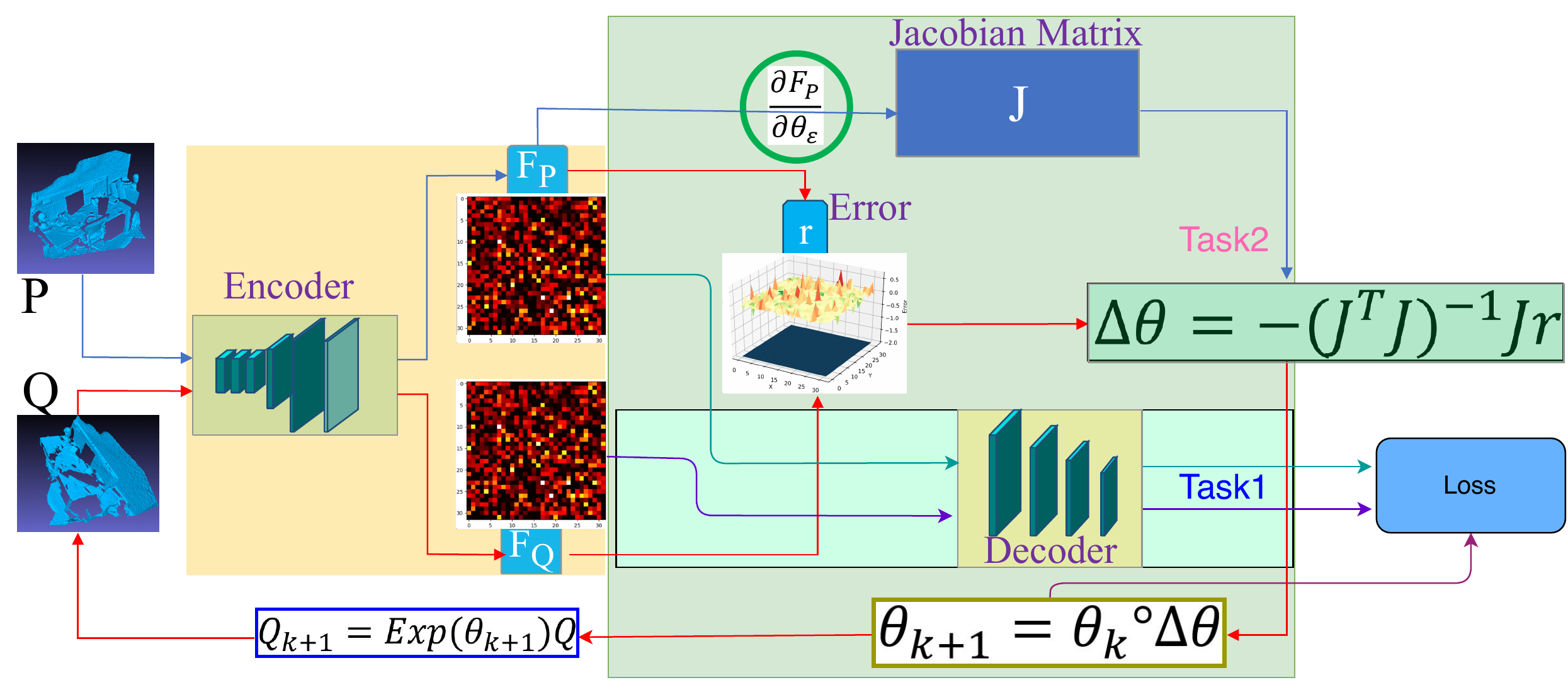}
	\caption{\textbf{High-level Overview of the proposed registration framework}. (1)- \textbf{\textcolor{red}{ Encoder}} extracts the features for two input point clouds (\textbf{P} and \textbf{Q}). (2)- Then, a multi-task semi-supervised neural network aims to solve the registration problem without correspondence. In the \textit{Task1}, \textbf{\textcolor{blue}{Decoder}} decodes the features. The whole encoder-decoder branch trains the encoder network in a unsupervised manner. In the \textit{Task2}, \textbf{\textcolor{green}{feature-metric projection error (\textbf{r})}} is calculated based on two input features ($F_P$ and $F_Q$). The feature error is input into a non-linear optimisation algorithm to estimate transformation increment ($ \bigtriangleup \theta$) and the transformation parameters are updated. Using the updated transformation parameters ($\theta_{k+1}$), the input point cloud Q is transformed and the whole process iteratively runs.}
	\label{lmnet}
\end{figure*}

\subsection{End-to-end learning-based registration methods}
The basic idea of end-to-end learning methods is to transform the registration problem into a regression problem. The recent methods \cite{Yang_2019_CVPR} tries to learn a feature from the point clouds to be aligned and then regresses the transformation parameters from the feature. \cite{wang2019non} proposes a registration network to formulate the correlation between source and target point sets and predicts the transformation by using the defined correlation. \cite{elbaz20173d} proposes an auto-encoder registration network for localisation, which combines super points extraction and unsupervised feature learning. \cite{lu2019deepicp} proposes a keypoint detection method and estimates the relative pose simultaneously. \cite{aoki2019pointnetlk} proposes a method to use the pointnet to solve the registration task, while its limitation is that category labels are required to initialize the network which is unrealistic in real application.  In a small summary, the current end-to-end learning methods regard transformation parameter estimation as a black box, and the distance metric is measured in the coordinate-based Euclidean space, which is sensitive to noise and density difference. Different from the present works, we propose a semi-supervised registration framework to learn to extract distinctive features and solve the transformation parameter estimation by minimising a feature-metric error without correspondences.  

In this paper, we investigate how to fused leverage both the classical non-linear optimisation theory and deep learning to directly estimate the transformation parameters without correspondences. In the end, a new semi-supervised feature-metric registration framework is proposed to customise the feature learning for the registration. This method is significantly different from the classical registration pipeline that alternatively solves the correspondence estimation and transformation parameter estimation. At the same time, instead of discarding the conventional mathematical theory and replacing the registration process by a whole black-box neural network, we would like to leverage both advantages by combing the classical registration mathematical theories with deep learning techniques. Furthermore, unlike conventional registration methods that minimise geometric projection error, our approach minimises feature-metric projection error.   

\section{Feature-metric Point Cloud Registration}
\label{algo-fmreg}
In this section, firstly, the problem formulation and overview of the registration framework are described. Secondly, we explain the \textit{Encoder} module in details. Thirdly, we show how to learn a distinctive feature and solve the registration in multi-task streams. Fourthly, the loss functions are detailed explained.  

\subsection{Problem Formulation}
Given two point clouds $P\in \mathbb{R}^{M\times 3}$ and $Q\in \mathbb{R}^{N\times 3}$, the goal of registration is to find the rigid transformation parameters $g$ (rotation matrix $R \in \mathcal{SO}(3)$ and translation vector $t \in \mathbb{R}^3$) which best aligns the point cloud $Q$ to $P$ as shown below:
\begin{equation}
\begin{split}
\operatorname*{arg\,min}_{R\in \mathcal{SO}(3), t\in \mathbb{R}^3} \| r(F(P),F(RQ+t)) \|_2^2 
\end{split}
\label{eq1} 
\end{equation}
where $r(F(P),F(RQ+t)) = \| F(P) - F(RQ+t)\|_2$ is the feature-metric projection error between $P$ and transformed $Q$. $F(P) \in \mathbb{R}^{K}$ is the feature of the point cloud $P$,  $K$ is the feature dimension (1024 is used in the experiments), $F$ is a feature extraction function learned by a \textit{Encoder} module.

To solve the above equation (\ref{eq1}), we propose a feature-metric registration framework to integrate both the merits of the classical non-linear algorithm and unsupervised learning techniques. The framework can be trained in a semi-supervised or unsupervised manner. Figure \ref{lmnet} shows an overview of the algorithm. Firstly, two rotation-attentive features are extracted for two input point clouds. Secondly, the features are input into a multi-task module. In the first branch (Task1), a decoder is designed to train the \textit{Encoder} module in an unsupervised manner. In the second branch, the projection error $r$ is calculated to indicate the difference between two input features, and the best transformation is estimated by minimizing the feature difference. The transformation estimation iteratively runs, and the transformation increment($\bigtriangleup \theta$) of each step is estimated by running the inverse compositional (IC) algorithm \cite{baker2004lucas}:
\begin{equation}
\bigtriangleup \theta = (J^TJ)^{-1}(J^Tr)
\end{equation}
where $r$ is the feature-metric projection error and $J=\frac{\partial r}{\partial \theta}$ is the Jacobian matrix of $r$ with respect to transformation parameters ($\theta$). Refer to \cite{baker2004lucas}, direct computation of $J$ in 2D images is extremely costly. The approach applied in the classical 2D images, the chain rule is utilised to estimate the Jacobian into two partial terms: the gradient of the image and the Jacobian of warp. However, this approach is not working for unordered 3D point clouds since there is not grid structure to allow us to compute the gradient of the 3D point cloud in $x$, $y$ and $z$ directions \cite{aoki2019pointnetlk}. 

To efficiently compute the Jacobian matrix, instead of calculating the Jacobian matrix by stochastic gradient approach, refer to \cite{aoki2019pointnetlk}, we utilise a different finite gradient to compute the Jacobian matrix:

\begin{equation}
J_i = \frac{\partial F_P}{\partial \theta_\xi} = \frac{F(R_iP + t_i)-F(P)}{\xi}
\end{equation}
where $\xi=(R_i,t_i)$ is the infinitesimal perturbations of the transformation parameters during the iteration.  In this work, six motion parameters are given for the perturbations that three angle parameters $(v_1, v_2, v_3)$ for the rotation and three jitter parameters $(t_1, t_2, t_3)$ for the translation. Refer to \cite{aoki2019pointnetlk}, the small fixed value ($2*e^{-2}$) on all the iteration generate the best results. After several iteration time (10 is used for the experiments), the proposed method outputs a final transformation matrix $g_{est}$ ($R$ and $t$) and feature-metric projection error $r_{est}$.

The critical components of the proposed framework are the \textit{Encoder} module and multi-task branches. We will now provide details about these components.

\subsection{Encoder }
The encoder module aims to learn a feature extraction function $F$, which can generate a distinctive feature for input point clouds. The main principle of designing the encoder network is that the generated feature should be rotation-attentive so that it can reflect the rotation difference during the transformation estimation. Refer to PointNet \cite{qi2017pointnet}, the feature is extracted by two \textit{mlp} layers and a max-pool layer. We discard the input transform and feature transform layers to make the feature aware of the rotation difference.

\subsection{Multi-task Branches}
After the features extracted, the next steps are feature learning and point cloud transformation estimation. The estimation is finalized by directly minimizing the feature-metric projection error and does not need search correspondences. This design will learn suitable features for iterative optimization-based methods on feature space.

\subsubsection{Encoder-Decoder Branch (Task1):}
After the encoder module generating distinctive features, inspired by \cite{groueix2018atlasnet}, we use a decoder module to recover the features back into 3D point clouds. This encoder-decoder branch can be trained in an unsupervised way which helps the \textit{Encoder} module to learn to generate distinctive feature being aware of rotation difference. For two rotated copies $PC1$ and $PC2$ of a point cloud, the principle of this branch is that the $Encoder$ module should generate different features for $P1$ and $P2$ and the decoder can recover the different features back to their correspondent rotated copies of the point cloud. This principle guarantees the success of unsupervised learning in training a distinctive feature extractor for the registration problem.

The decoder block consists of four-layer fully-connected layers and activates by LeakyReLU. The output of decoder module is the same dimension to the input point cloud.

\subsubsection{Feature-metric Registration Branch (Task2):}
To solve the registration problem, as discussed in the \textit{problem formulation} section and showed in Figure \ref{lmnet}, we estimate the transformation parameters by using the inverse compositional algorithm (non-linear optimisation) to minimize a feature-metric projection error. The feature-metric projection error is defined as 

\begin{equation}
r = \| F(P) - F(g\cdot Q)\|_2^2
\end{equation}
where $F(.)\in \mathbb{R}^{K}$ is a global feature of point cloud ($P$ or $g\cdot Q$), and $g$ is the transformation matrix ($R$ and $t$).

\begin{figure*}[ht]
	\includegraphics[width=\linewidth,height=5cm]{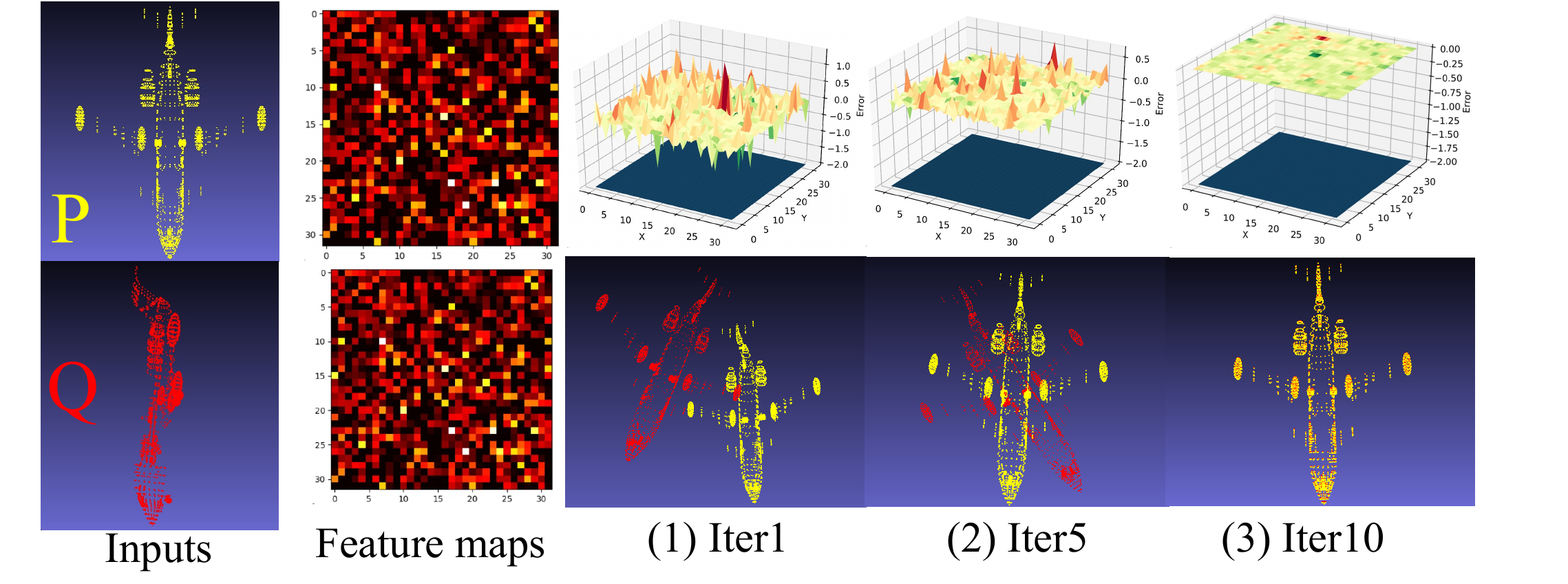}
	\caption{{The feature maps generated by the feature extraction network and the feature difference during the iteration process.} At the first iteration, the feature map difference (top) is large between $P$ and transformed $Q$, and their alignment (bottom) is not well. At the $10^{th}$ iteration, the feature map difference of point cloud $P$ and transformed $Q$ becomes extremely small and the alignment is almost perfect. }
	\label{feat}
\end{figure*}
To have a better intuition about what does the feature network extracted during the non-linear optimisation process, we visualise the feature maps and the iterative process in Figure \ref{feat}. We reshape the global feature maps into a square matrix and show it into as an image. Figure \ref{feat} shows the feature maps of point cloud $P$ and transformed point cloud $Q$, and feature map differences in the first iteration, $5^{th}$ and the final $10^{th}$ iteration. Looking at the last three columns of Figure \ref{feat}, the feature map difference (top row) becomes smaller when the alignment becomes more accurate (bottom row).

\subsection{Loss functions}
The key task for training is the \textit{Encoder} module which used to extract a rotation-attentive feature. We propose two loss functions to provide a semi-supervised framework. It can be easily extended to unsupervised framework by directly ignoring the supervised geometric loss.

\subsubsection{Chamfer loss}
The encoder-decoder branch can be trained in an unsupervised manner. Refer to \cite{groueix2018atlasnet}, Chamfer distance loss is used:
\begin{eqnarray}
loss_{cf} = \sum_{p\in A}^{}\sum_{i=1}^{N} \min_{q\in S^*}\|\phi_{\theta_i}(p;x)-q\|^2_2 \\
+ \sum_{q\in S^*}^{}\min_{i,in{1..N}}\min_{p\in A}\|\phi_{\theta_i}(p;x)-q\|^2_2 \nonumber
\end{eqnarray}
where $p\in A$ is a set of points sampled from a unit square $[0,1]^2$, $x$ is a point cloud feature,  $\phi_{\theta_i}$ is the $i^{th}$ component in the MLP parameters, $ S^*$ is the original input 3D points.

%

\begin{figure*}[ht] 
	\centering	
	\includegraphics[width=0.47\textwidth,height=3.4cm]{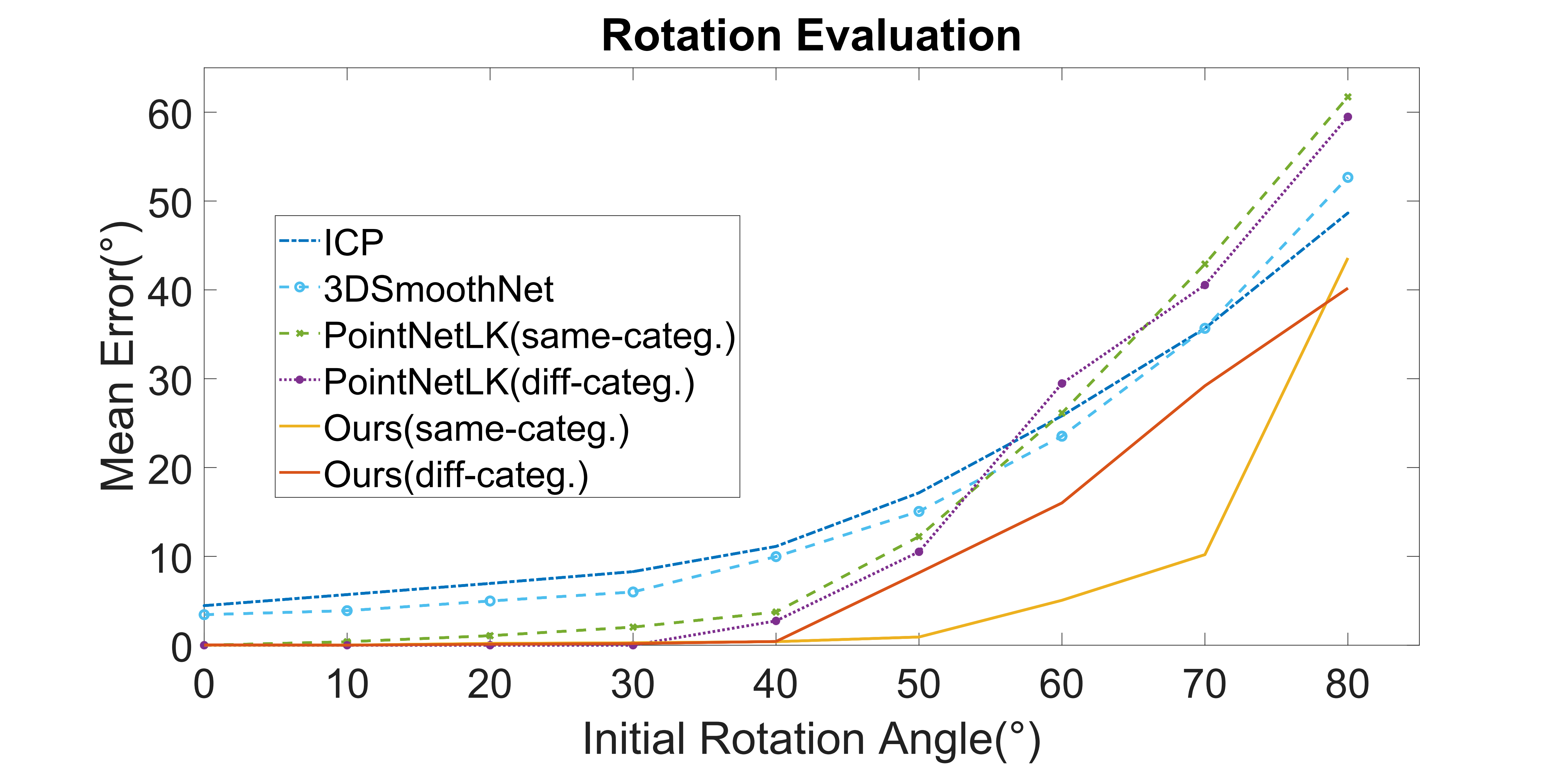} 
	\includegraphics[width=0.47\textwidth,height=3.4cm]{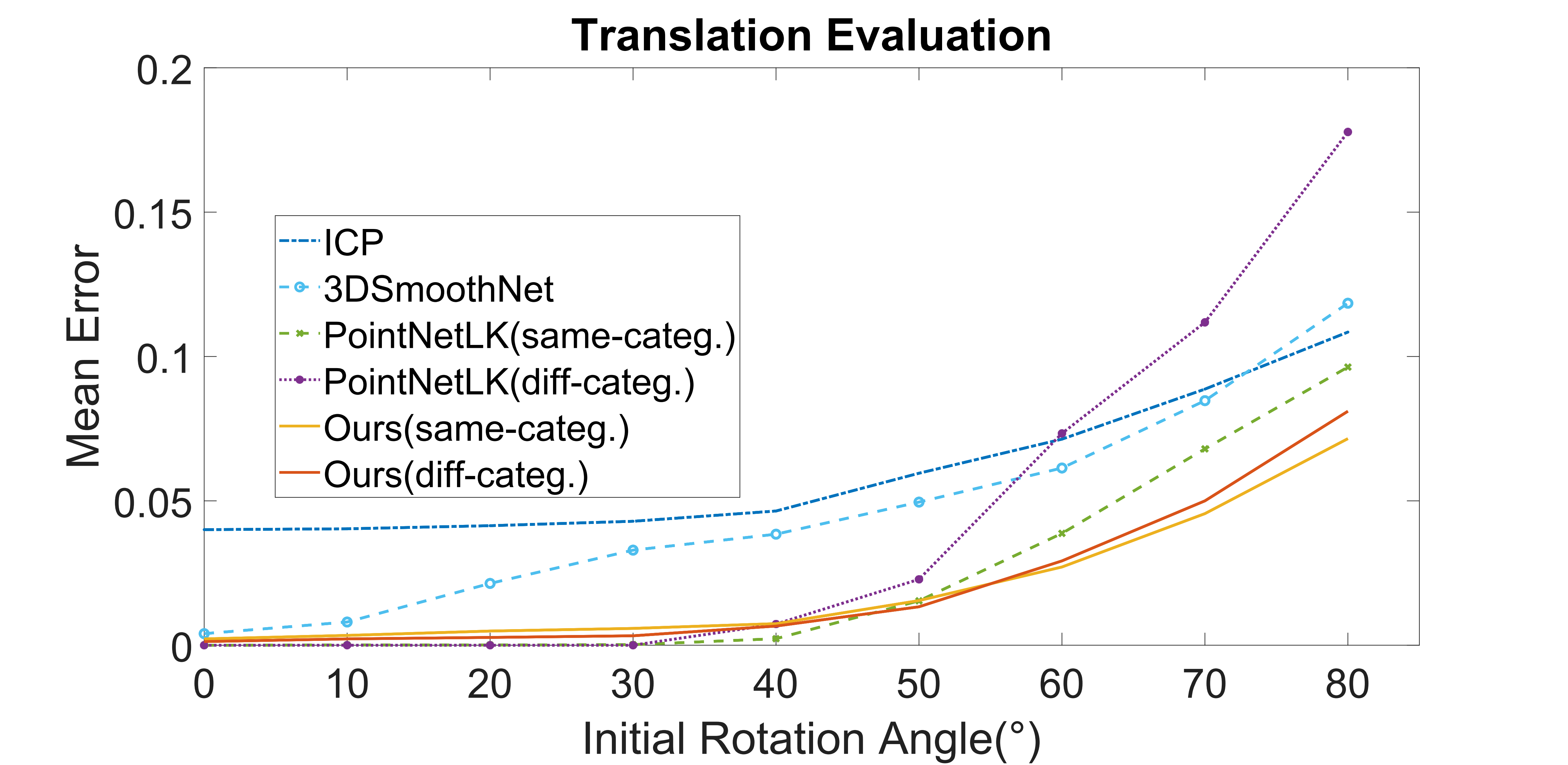}
	\caption{Comparison of the same and different categories under different angles' rotation. The results show our method obtains better accuracy than the classic optimisation method, feature learning method and deep learning registration method.}
	\label{frot_trans}
\end{figure*}
\subsubsection{Geometric loss}
The goal of the geometric loss function is to minimize the difference between the estimated transformation matrix ($g_{est}$) and the ground-truth transformation matrix ($g_{gt}$). To balance the influence of translation and rotation, inspired by \cite{lv2019taking} in solving 2D image alignment, we exploit the 3D point error (PE) between the different transformation of $g_{est}$ and $g_{gt}$. Therefore,  the loss function is 
\begin{equation}
loss_{pe} =	\frac{1}{M}\sum_{i=1}^M \left\| f(g_{est}\cdot P) - f(g_{gt}\cdot P)\right\|_2^2
\end{equation}
where $P$ is a point cloud, and $M$ is its total point number.

The final loss function of semi-supervised training is:
\begin{equation}
loss = loss_{cf} + loss_{pe} 
\end{equation}
For unsupervised training, we only use $loss_{cf}$.

\section{Experiments}

\subsection{Datasets}

\textbf{ModelNet40} To systematically evaluate highly varying object motions, we use ModelNet40 dataset. The ModelNet40 \cite{wu20153d} contains 3D CAD models from 40 categories, which is a widely used dataset to train 3D deep learning network. The dataset is split into two parts: 20 categories for training and same-category testing, and another 20 categories for cross-category testing. In the same-category experiments, we split the dataset into $8:2$, where $80\%$ are used for training and $20\%$ for testing. The source point cloud is the vertices from ModelNet40, which is normalized into a unit box at the origin $[0, 1]^3$. Then, the target point cloud is generated by applying a rigid transformation to the source point cloud. The registration performance is evaluated and compared by using the source and target point clouds. During the training, the rigid transformation $T_{gt}$ is randomly generated, where the rotation is in the range $[0, 45]$ degrees with arbitrarily chosen axes and translation is in the range $[0, 0.8]$. For a fair comparison, initial translations for testing are in the range $[0, 0.8]$ and initial rotations are in the range $[0, 80^\circ]$.

\textbf{7Scene \cite{shotton2013scene}} is a widely used benchmark registration dataset by a Kinect RGB-D camera on the indoor environment. It contains seven scenes including Chess, Fires, Heads, Office, Pumpkin, Redkitchen and Stairs. Following 3DMatch \cite{3dmatch}, we project the depth images into point clouds and fuse multi-frame depth thorough truncated signed distance function(TSDF) fusion. The dataset is divided into 296 scans for training and 57 scans for testing. The rotation is initialized in the range of $[0, 60^\circ]$ and the translation is initialized in the range of $[0, 1.0]$ both for training and testing. 

\subsection{Baselines}

We implemented the following baselines:

(1) \textbf{LM-ICP} \cite{fitzgibbon2003robust}: To compare with the classic registration method, we use the Levenberg–Marquardt (LM)-based iterative closest point (LM-ICP). 

(2) \textbf{3DSmoothNet + RANSAC} \cite{gojcic2019perfect}: To compare with the feature learning methods, which combine a learned feature and a classical registration method, we select the state-of-the-art 3DSmoothNet +RANSAC as our comparison method. In the comparison experiments, 5K points are sampled.

(3) \textbf{PointNetLK} \cite{aoki2019pointnetlk}: To compare with the registration method of using deep learning, we use PointnetLK, which combines Pointnet with Lucas-Kanade algorithm. We train the PointnetLK using the same way to our method.

\textbf{Evaluation Criteria}. Following the \cite{aoki2019pointnetlk}\cite{pais20193dregnet}, the evaluation is performed by calculating the difference of the estimated  transformation parameters ($g_{est}$) against the ground truth ($g_{gt}$). The angular is calculated as $\theta = log(R)$ and the angular error can be calculated as Euclidean distance between $\theta_{est}$ and $\theta_{gt}$. The translational error is calculated as the Euclidean distance between $t_{est}$ and $t_{gt}$. The $RMSE$ represents the root-mean-square error between $g_{est}$ and $g_{gt}$.

\subsection{Evaluation on ModelNet40. }
For network training, PointNetLK initializes the feature extraction network by a standard classification task on ModelNet40 and fine-tune the neural network for 400 epochs. 3DSmoothNet initializes the network using 16-bin pre-trained model and fine-tunes on the training dataset of ModelNet40. Our method does not rely on any classification labels and trains in an end-end stream. The method is implemented in PyTorch and trained on an NVIDIA P5000.

\textbf{Comparison Results}: Figure \ref{frot_trans} summarizes the comparison results on the dataset with different initial rotation angles and random translation. Compared to all baseline methods, our method achieves the overall best performance across different initial angles. Especially, the proposed method shows tiny registration error before $60^\circ$, which is less than $0.02$. 

Looking deeper into the experimental results in Figure \ref{frot_trans}, the classical ICP is sensitive to the initial angles so that the error increase when the initial angles grow. Interestingly, although the 3DSmoothNet learns a rotation-invariant descriptor, the registration error still goes up when the initial angle increases. The interpretation contains two folds. 1) The feature learning is separated from the registration goal, and the learned model is only suitable for point cloud matching but not perfect for the registration task. 2) The data in our experiments use the vertexes directly from the CAD models, which contain a lot of similar local structures — the local feature learning for matching faces an ambiguous problem.  Instead of learning indirect features for matching, our method solves the registration problem by directly learning a rotation-attentive feature for point cloud registration. Besides, our method obtains apparent better performance than PointNetLK in large angles ($30^\circ-80^\circ$). The interpretation is that our semi-supervised solution can generate much better features, which can estimate a better registration solution.

\begin{figure*}[ht] 
	\centering	
	\includegraphics[width=0.47\textwidth,height=3.3cm]{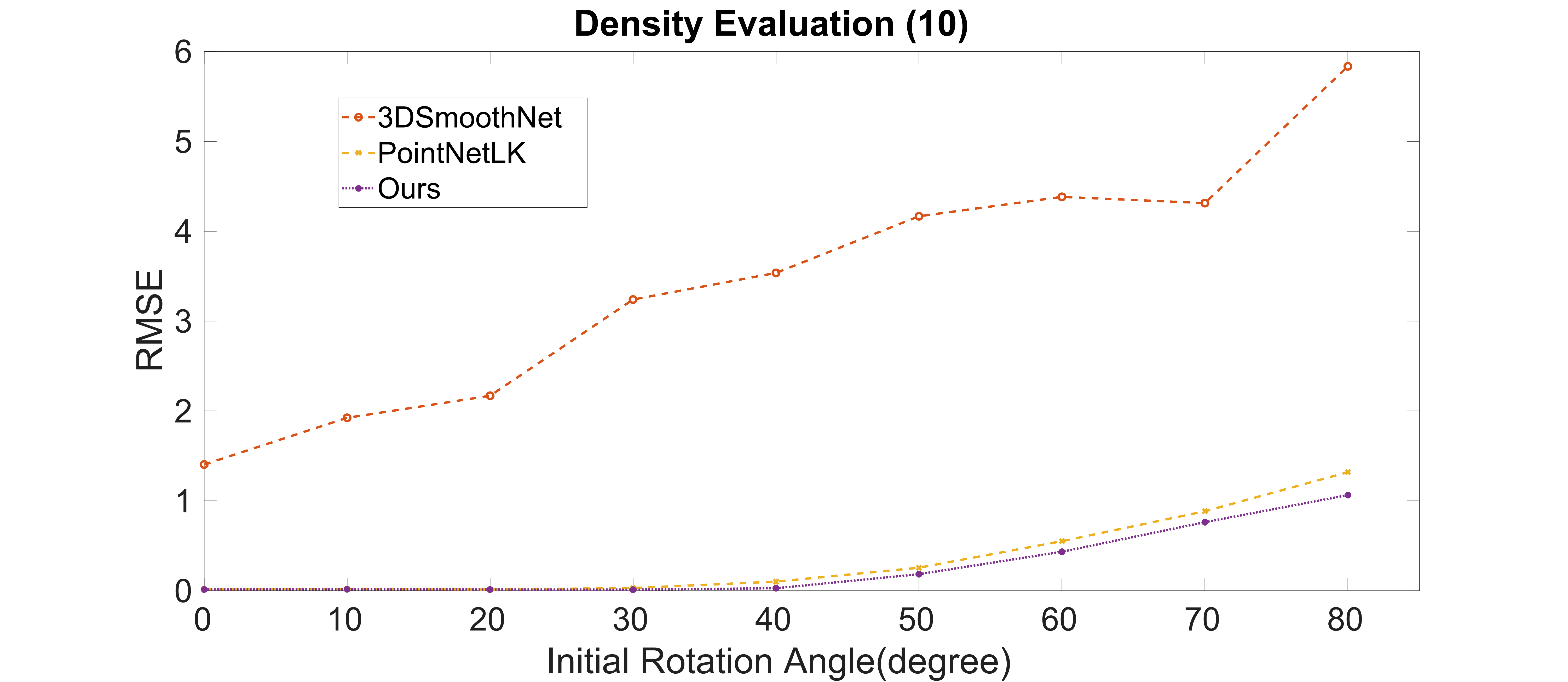}
	\includegraphics[width=0.47\textwidth,height=3.3cm]{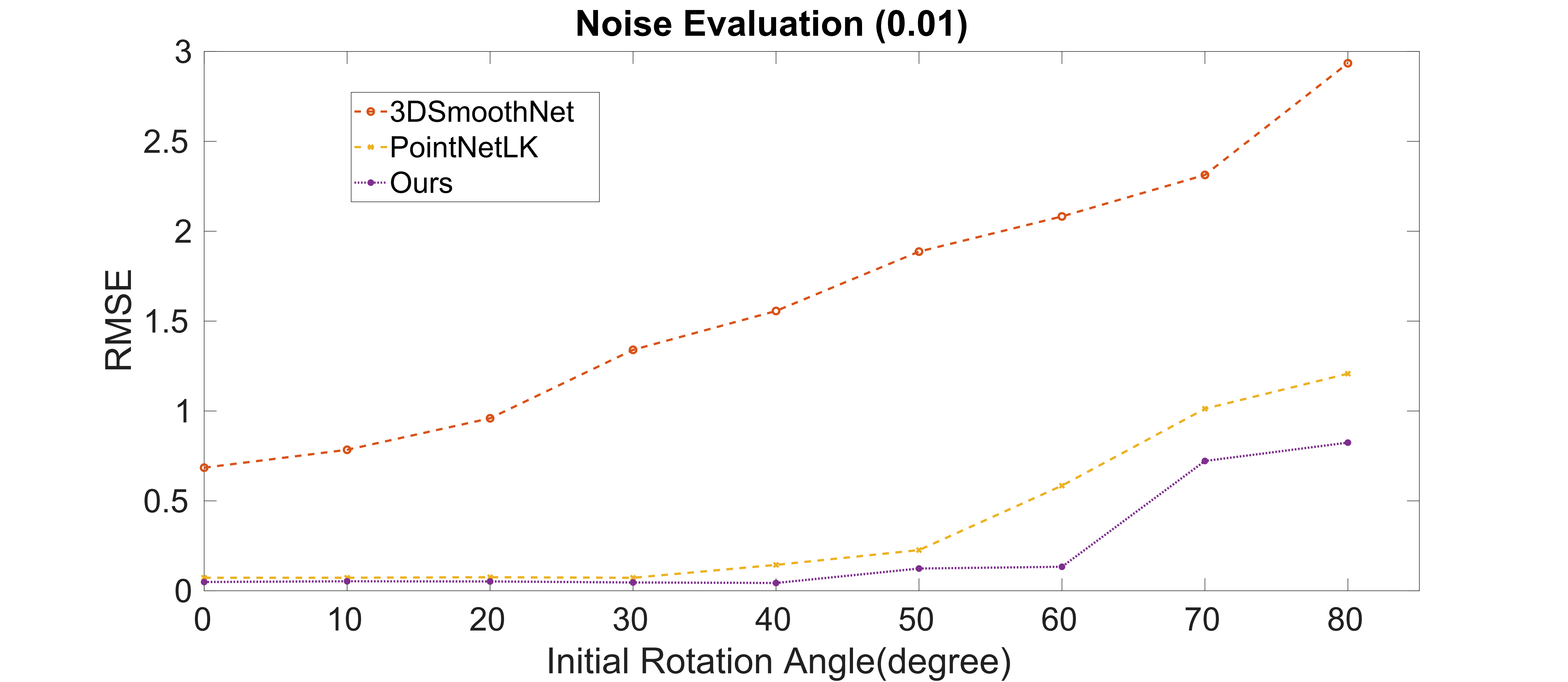} 
	\caption{Comparison results with the impact of density difference and noise. The left column shows the comparison results on 10 times density difference. The right column shows the comparison results under Gaussian noise with signal-to-noise ratio(SNR) 0.01.}
	\label{density}
\end{figure*}
\begin{figure}[ht]
	\includegraphics[width=\linewidth,height=4cm]{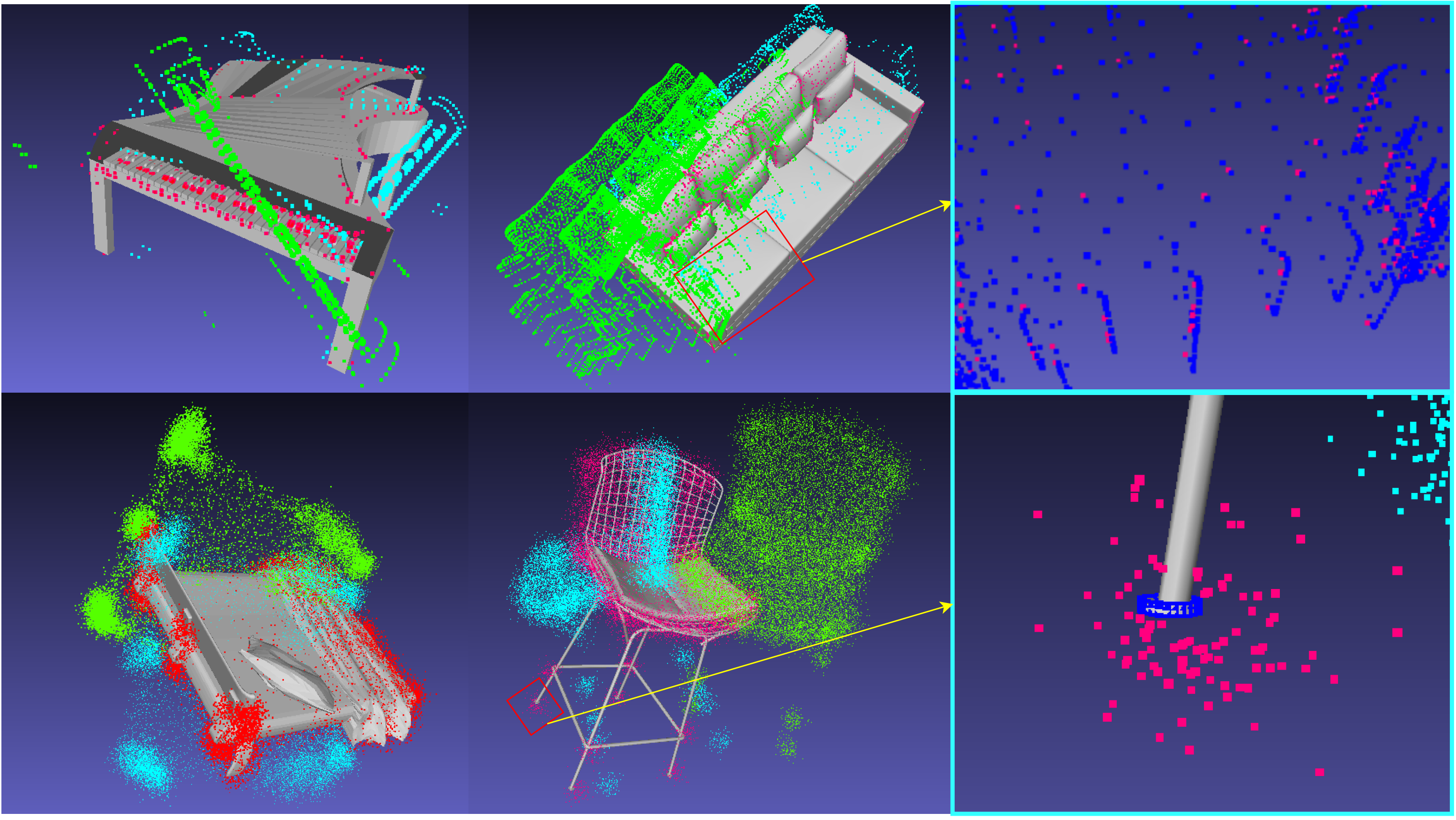}
	\caption{Ablation study on high density difference (Top) and high nosie (bottom). Our method (red) shows obviously better accuracy than 3DSmoothNet (green) and PointnetLK (light blue). The blue points in the last column are the original points without variants.}
	\label{nosie_outlier}
\end{figure}

\textbf{Density Difference}: To demonstrate our method can handle significant density difference, one of the original point cloud (source) was randomly removed $90\%$ points to generate ten times' density difference. The left column of Figure \ref{density} shows the results of density difference, which demonstrates that the proposed method achieves better registration results in high-density difference. To visually show the results, we select the initial rotation angle as $60^\circ$. The top row of Figure \ref{nosie_outlier} shows that the proposed method can align accurately on the large density difference (See the red points align well with the CAD model). However, 3DSmoothNet (green points) and PointNetLK(light blue) face challenging on aligning these point clouds with a high-density difference. The third column of the first row in Figure \ref{nosie_outlier} shows the challenging density difference case.

\textbf{Large Noise}: We also conduct experiments on noisy data. The right column of Figure \ref{density} shows the registration results with Gaussian noise ratio. Although the point clouds are impacted by different noise, our method obtains the lowest root-mean-square error of transformation matrix difference \cite{evangelidis2017joint}, which demonstrates the proposed method always obtain high robustness under noise variation.   
The bottom row of Figure \ref{nosie_outlier} shows a visual comparison results where the data is impacted by Gaussian noise with a ratio of $0.02$. The result shows the proposed method can align accurately on the noisy data. The right column of the bottom row in Figure \ref{nosie_outlier} demonstrates that the noise is large and challenging. Our method can overcome this large noise while the compared methods fail.

\begin{figure}[ht] 
	\centering	
	\includegraphics[width=0.47\textwidth]{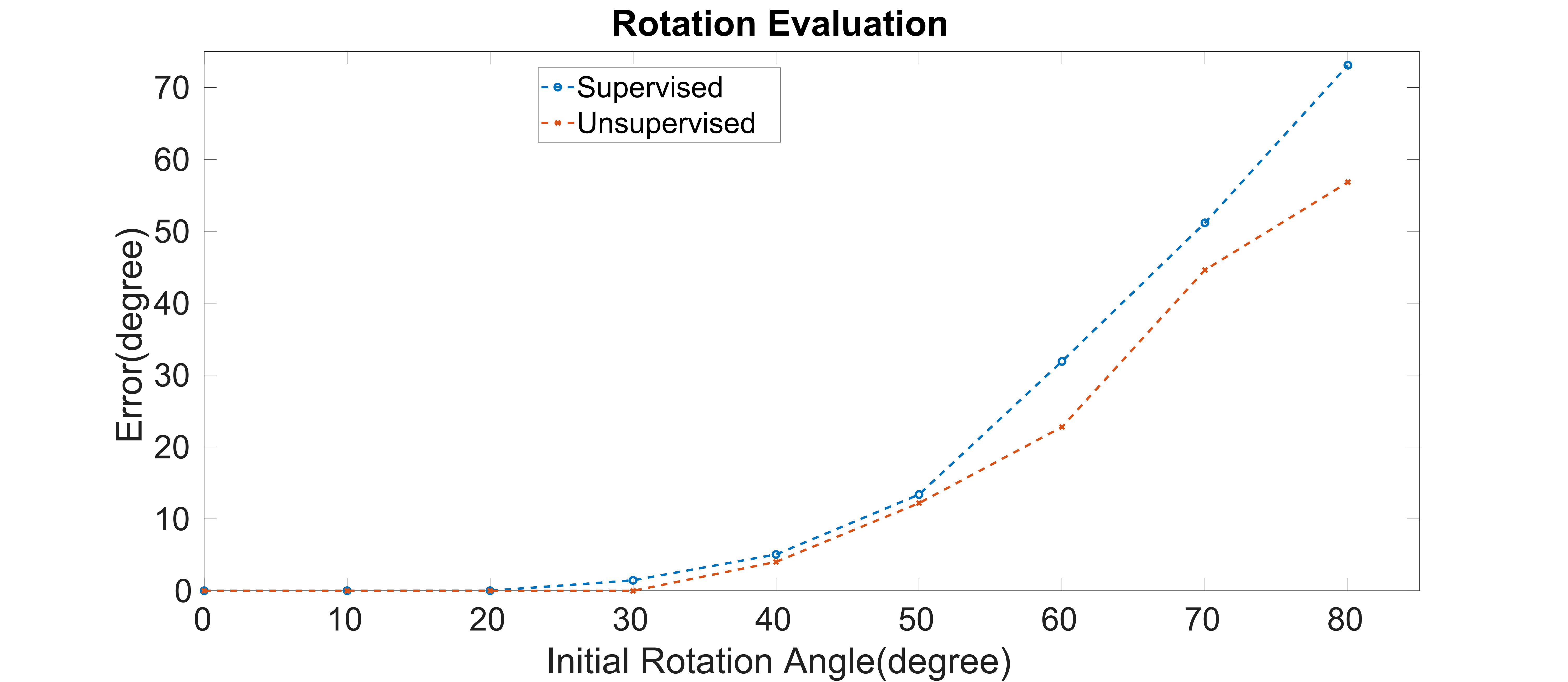} 
	\caption{Comparison of training the model in a semi-supervised or unsupervised way. The results show that the unsupervised training obtains better performance.}
	\label{unsupervised}
\end{figure}
\textbf{Semi-supervised \textit{VS} Unsupervised}
To investigate the performance of the unsupervised encoder-decoder branch, we train the \textit{Encoder} module using by only considering the \textit{Chamfer loss}. Then, both semi-supervised and unsupervised are evaluated on the ModelNet40 with the same experiment set before. Figure \ref{unsupervised} shows that the unsupervised method can always obtain better results than the semi-supervised method. These experimental results show the proposed feature-metric registration framework can have a wide application since no labels are required to train the neural network. In this paper, to make a fair comparison with PointNetLK, most experiments are conducted in a semi-supervised manner. The better results also demonstrate the unsupervised branch improve the registration robustness and accuracy.

\begin{figure*}[ht]
	\includegraphics[width=\linewidth,height=2.5cm]{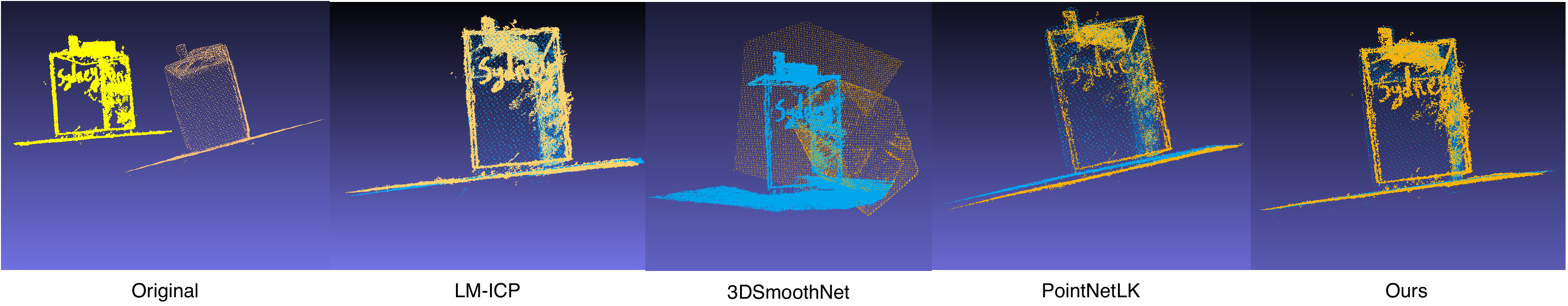}
	\caption{Visually comparison results in solving cross-source point clouds registration. }
	\label{cross-source}
\end{figure*}
\begin{figure*}[ht]
	\includegraphics[width=\linewidth, height=2.5cm]{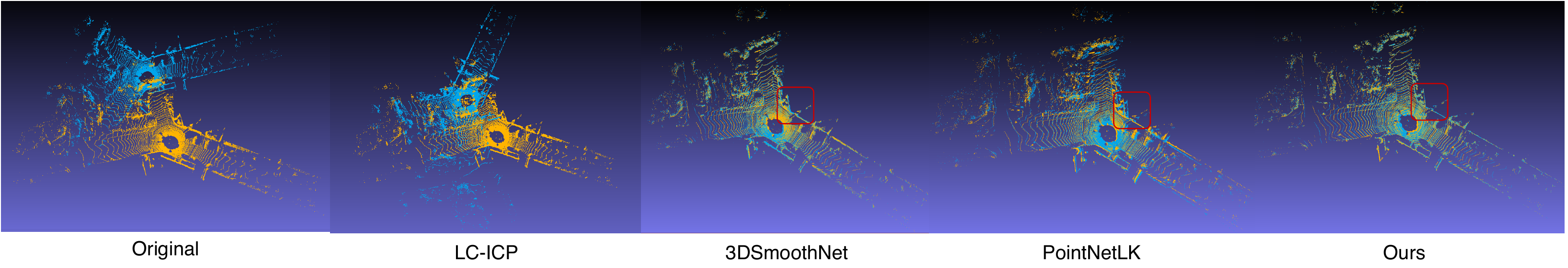}
	\caption{Visually comparison results in solving sparse outdoor Lidar point clouds registration.}
	\label{lidar}
\end{figure*}

\subsection{Evaluation on 7Scenes}
 Since there are no category labels available in 7Scenes dataset, we directly train the PointNetLK without feature network initialization.  Our method is trained directly in an unsupervised manner. The pre-trained model of 16 bin is selected for 3DSmoothNet since this model is already trained on 3DMatch\cite{3dmatch} dataset, which includes \textit{7Scene} dataset.  We evaluate the rotation angle and translation difference between algorithm estimation and groundtruth and compare the registration performance with state-of-the-art methods. 

\textbf{Comparison Results}: Table \ref{tableAngle} summarizes the performance comparison results on 7Scene. Our method outperforms the classic optimization method, feature extraction method and the recent deep learning method. We believe this is due to the unsupervised branch can train the feature extraction network to generate a distinctive feature to understand the point cloud, and the feature-metric registration framework provides an effective way to solve the registration problem. 
\begin{table}[h]
	\begin{center}
		\begin{tabular}{p{1.0cm}|p{1.2cm}p{1.0cm}p{1.0cm}p{0.8cm}p{0.8cm}}
			\hline
			Method & LM-ICP &\cite{gojcic2019perfect}  & \cite{aoki2019pointnetlk}& Ours\\
			\hline
			Rot.($^\circ$) &  12.7 &1.2  & 15.3 & 0.004 \\
			Trans.& 0.96 &0.05  & 0.92 & 0.002\\
			\hline
		\end{tabular}
	\end{center}
	\caption{Quantitative comparisons on the 7Scene dataset with the classical optimisation method (LM-ICP), learned feature method \cite{gojcic2019perfect}(3DSmoothNet) and recent deep learning method \cite{aoki2019pointnetlk}(PointNetLK)}
	\label{tableAngle}
\end{table}

\begin{figure}[ht]
	\includegraphics[width=\linewidth,height=3cm]{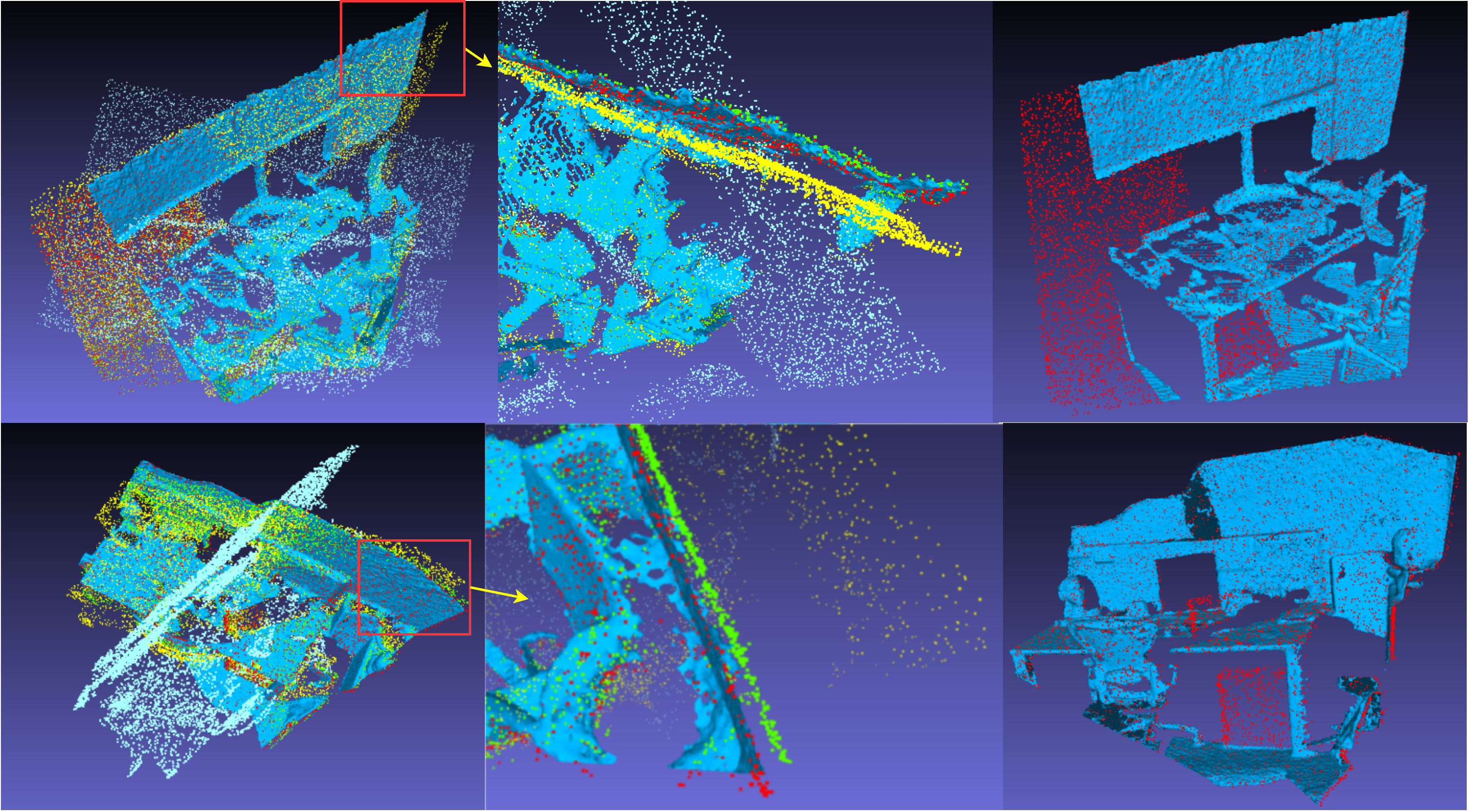}
	\caption{Ablation study on partial overlap. Our method (red) shows obviously better accuracy than LM-ICP (yellow), 3DSmoothNet (green) and PointnetLK (light blue). The last column shows the partial overlap.}
	\label{partialvisual}
\end{figure}

\textbf{Partial overlap}:
To compare the performance on the partial overlap, we manually remove part of the scan (top row of Figure \ref{partialvisual}) or some objects inside the scans (bottom row of Figure \ref{partialvisual}). Figure \ref{partialvisual} shows that the proposed method can handle partial overlap problem, and obtain better performance.

\textbf{Generalize to cross sources}:
Recently, the 3D sensor hardware and 3D reconstruction software have undergone tremendous changes. Cross-source point clouds are data from different types of image sensors. As 3D data become widely existed, data fusion of cross-source point clouds can leverage different benefits in different image sensors. According to \cite{huang2016coarse, huang2017systematic,huang2017coarse}, cross-source point cloud registration is challenging since the data contain large variations such as large noise and outliers, density difference. To demonstrate the performance of our method in solving the cross-source point clouds, following \cite{huang2017systematic}, we capture cross-source point clouds by using Kinect and RGB camera + 3D reconstruction software and remove the scale variation. Figure \ref{cross-source} shows our method can also handle the cross-source point cloud registration and obtain a better result than the recent state-of-the-art registration methods.

\textbf{Generalize to outdoor point clouds:}
Figure \ref{lidar} shows results on sparse outdoor Lidar point cloud registration using KITTI point cloud sequences. Our method obtains better results than other methods as shown in the redbox.

\subsection{Time complexity}
We also run the running time comparison on point clouds with around 12K points. Table \ref{time} shows that our method achieves eight times faster than LM-ICP, seven times faster than 3DSmoothNet and two times faster than PointNetLK. We find that our method can usually achieve the best alignment within five iterations.

\begin{table}[h]
	\begin{center}
		\begin{tabular}{p{1.5cm}|p{1.2cm}p{1.8cm}p{1.5cm}p{0.5cm}}
			\hline
			Method & LM-ICP & 3DSmoothNet & PointNetLK& Ours\\
			\hline
			Time(s)  & 10.8  & 9.6 & 2.4 & 1.3  \\
			\hline
		\end{tabular}
	\end{center}
	\caption{Running time comparison on point clouds with around 12K points.}
	\label{time}
\end{table}

\section{Conclusion}
In this paper, we propose a feature-metric framework to solve the point cloud registration, and the framework can be trained using a semi-supervised or unsupervised manner. Experiments demonstrate that the proposed method achieves better accuracy than the classical registration method, state-of-the-art feature learning method and deep learning registration method. Our ablation studies show that the proposed method can handle significant noise, density difference and partial overlap. Besides, experiments show that our method can also achieve robust, accurate and fast registration by training in an unsupervised manner. We believe the feature-metric framework will provide high potential for both academic and industrial applications. 

{\small
\bibliographystyle{ieee_fullname}
\bibliography{egbib}
}

\end{document}